\documentclass[conference]{IEEEtran}
\IEEEoverridecommandlockouts

\usepackage[backend=biber, style=ieee, url=false]{biblatex}
\usepackage{amsmath,amssymb,amsfonts}
\usepackage{algorithm}
\usepackage{algpseudocode}
\usepackage{graphicx}
\usepackage{textcomp}
\usepackage{xcolor}
\usepackage{booktabs}
\usepackage{pifont}
\usepackage{multirow}
\usepackage{boldline}
\usepackage{colortbl}
\usepackage{multirow}
\usepackage{hyperref}
\usepackage{caption}
\PassOptionsToPackage{table}{xcolor}
\usepackage{xcolor}
\usepackage{color}

\makeatletter
\newcommand{\groupedRowColors}[5][0]{
    \global\rownum=\z@
    \global\@rowcolorstrue
    \@ifxempty{#4}%
        {\def\@oddrowcolor{\@norowcolor}}%
        {\def\@oddrowcolor{\gdef\CT@row@color{\CT@color{#4}}}}%
    \@ifxempty{#5}%
        {\def\@evenrowcolor{\@norowcolor}}%
        {\def\@evenrowcolor{\gdef\CT@row@color{\CT@color{#5}}}}%
    \def\@rowcolors{%
        \if@rowcolors
            \noalign{%
                \relax
                \ifnum\rownum<#3
                    \@norowcolor
                \else \ifodd \numexpr (\rownum-#1)/#2\relax
                    \@oddrowcolor
                \else
                    \@evenrowcolor
                \fi \fi
            }%
        \fi
    }%
    \CT@everycr{\@rowc@lors\the\everycr}%
    \ignorespaces
}
\makeatother

\def\BibTeX{{\rm B\kern-.05em{\sc i\kern-.025em b}\kern-.08em
    T\kern-.1667em\lower.7ex\hbox{E}\kern-.125emX}}

\newcolumntype{C}[1]{>{\centering\arraybackslash}p{#1}}
\newcolumntype{L}[1]{>{\raggedright\arraybackslash}p{#1}}
\newcolumntype{R}[1]{>{\raggedleft\arraybackslash}p{#1}}

\newcommand{\signo}{\ding{56}}
\newcommand{\sigyes}{\ding{52}}

\newcommand{\NAME}{{AI-GenBench}}

\begin{document}

\title{\NAME: A New Ongoing Benchmark for AI-Generated Image Detection}

\author{
    Lorenzo Pellegrini\IEEEauthorrefmark{1}\IEEEauthorrefmark{4}\thanks{\IEEEauthorrefmark{4}Corresponding author: Lorenzo Pellegrini (l.pellegrini@unibo.it)}, Davide Cozzolino\IEEEauthorrefmark{2}, Serafino Pandolfini\IEEEauthorrefmark{1}, Davide Maltoni\IEEEauthorrefmark{1},\\ Matteo Ferrara\IEEEauthorrefmark{1}, Luisa Verdoliva\IEEEauthorrefmark{2}, Marco Prati\IEEEauthorrefmark{3}, Marco Ramilli\IEEEauthorrefmark{3}\vspace{0.20cm} \\
    
    \IEEEauthorrefmark{1}\textit{Dipartimento di Informatica - Scienza e Ingegneria (DISI)}\\\textit{Università di Bologna, Cesena, Italy}\vspace{0.22cm}\\ 
    \IEEEauthorrefmark{2}\textit{Dipartimento di Ingegneria Elettrica e delle Tecnologie dell'Informazione (DIETI)}\\\textit{Università degli Studi di Napoli Federico II, Naples, Italy}\vspace{0.22cm}\\
    \IEEEauthorrefmark{3}\textit{IdentifAI, Italy}
}

\IEEEaftertitletext{
\vspace{-0.8cm}
    \centering
    \includegraphics[width=0.9\linewidth, page=1, trim=0 300 0 -20]{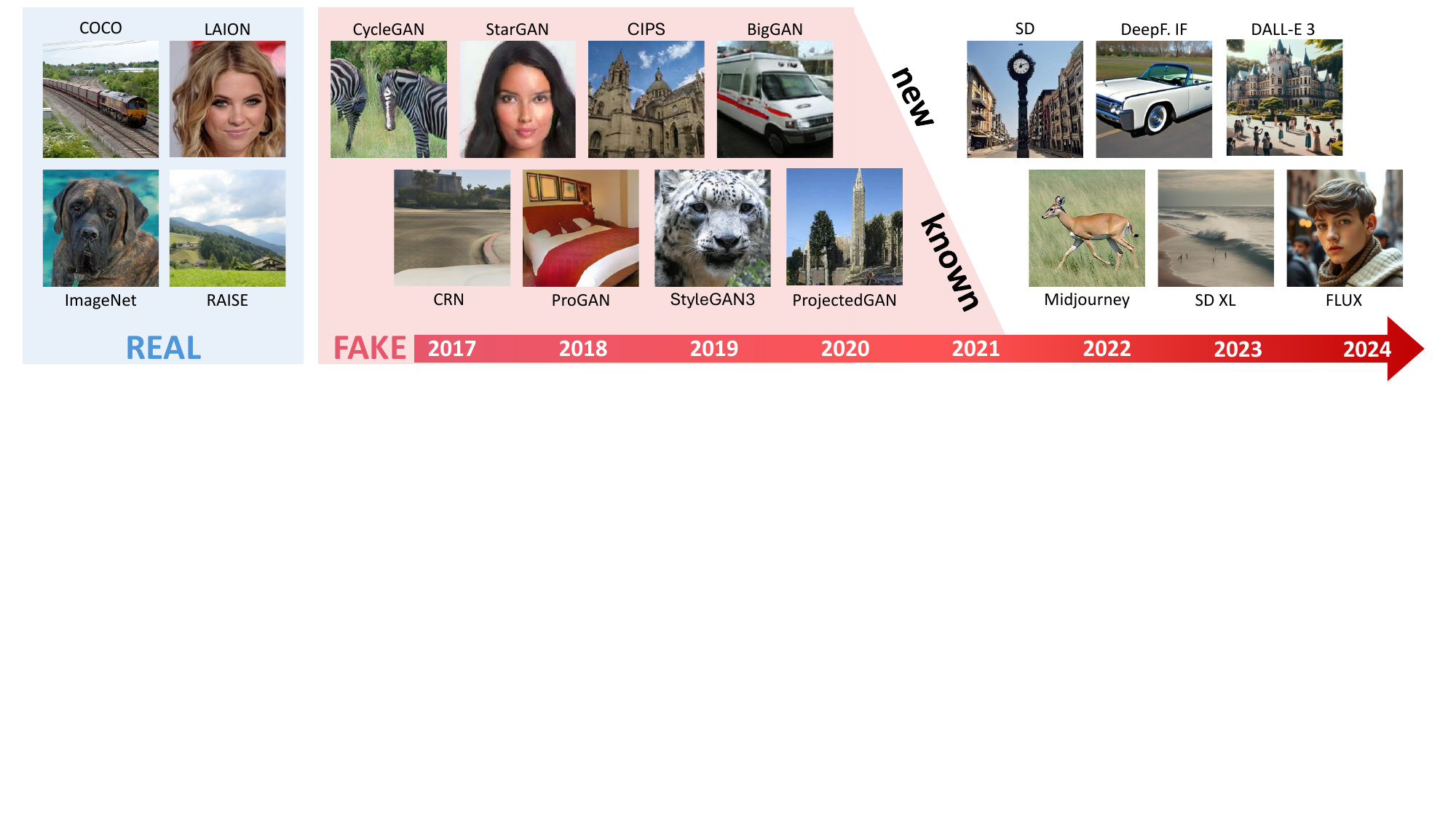} 
    \vspace{1mm}
    \captionof{figure}{We propose \NAME, an on-going benchmark for detecting AI-generated images in the wild. It introduces a temporal evaluation framework: training detectors on known generators and testing them on new ones as it happens in a realistic scenario. It also addresses critical limitations of current benchmarks and represents a valuable tool for researchers and fact-checkers. The benchmark comprises images generated by 36 different generators. 
}
    \label{fig:teaser}
\vspace{0.25cm}
}

\IEEEaftertitletext{%
\vspace{-0.2cm}
\noindent
\footnotesize
© 2025 IEEE. This is the author’s accepted version of the work.  
The final published version is available in the IEEE Xplore Digital Library:  
\url{https://doi.org/10.1109/IJCNN64981.2025.11228377}
\vspace{0.3cm}
}


\maketitle

\begin{abstract}
The rapid advancement of generative AI has revolutionized image creation, enabling high-quality synthesis from text prompts while raising critical challenges for media authenticity. We present \NAME, a novel benchmark designed to address the urgent need for robust detection of AI-generated images in real-world scenarios. Unlike existing solutions that evaluate models on static datasets, \NAME~introduces a temporal evaluation framework where detection methods are incrementally trained on synthetic images, historically ordered by their generative models, to test their ability to generalize to new generative models, such as the transition from GANs to diffusion models. Our benchmark focuses on high-quality, diverse visual content and overcomes key limitations of current approaches, including arbitrary dataset splits, unfair comparisons, and excessive computational demands. \NAME~provides a comprehensive dataset, a standardized evaluation protocol, and accessible tools for both researchers and non-experts (e.g., journalists, fact-checkers), ensuring reproducibility while maintaining practical training requirements. By establishing clear evaluation rules and controlled augmentation strategies, \NAME~enables meaningful comparison of detection methods and scalable solutions. Code and data are publicly available to ensure reproducibility and to support the development of robust forensic detectors to keep pace with the rise of new synthetic generators\footnote{\url{https://github.com/MI-BioLab/AI-GenBench}.}
\end{abstract}


\section{Introduction}
In recent years, the field of image generation has experienced rapid progress, marked by the development of robust and flexible tools based on diffusion models. These tools are capable of producing high-quality images from general conditional inputs, such as text, enabling professionals to leverage AI for innovative and creative applications in design, marketing, and entertainment. However, the potential misuse of these technologies raises significant ethical and social concerns, including the dissemination of disinformation and infringements of intellectual property rights \cite{epstein2023art, barrett2024identifying, lin2024detecting}. Consequently, there is an urgent need to develop effective methods for distinguishing real images from those generated by AI.

\begin{table*}[t!]
    \centering
    \caption{SoTA Forensic Benchmarks.}
    \resizebox{1.0\linewidth}{!}{\small
    \begin{tabular}{clccccccc}
    \toprule
        \textbf{Year} & \textbf{Acronym} & \textbf{Content} & \multicolumn{2}{c}{\textbf{Test data}}      & \multicolumn{2}{c}{\textbf{Training data}} & \textbf{Available} & \textbf{Temporal} \\
                      &                  &                  & \textbf{\#Real / \#Fake} & \textbf{\#gen.}  & \textbf{\#Real / \#Fake} & \textbf{\#gen.} & \textbf{Online}    & \textbf{ordering} \\
    \midrule
    2021 & \cite{he2021forgerynet}       ForgeryNet        & Face     & 290K / 290K & 15 & 1.2M / 1.2M  & 15 & \sigyes & \signo  \\
    2023 & \cite{wang2023benchmarking}   DeepArt           & Artworks & 2.3K / 2.4K &  5 &  62K / 71K   & 5  & \signo  & \signo  \\
    2023 & \cite{epstein2023online}      Epstein2023       & General  & ~22K / 116K & 14 & 202K / 454K  & 14 & \signo  & \sigyes \\
    2024 &  \cite{schinas2024sidbench}   SIDBench          & General  &  46K / 52K  & 16 & 360K / 360K  & 1  & \sigyes & \signo  \\
    2024 & \cite{chen2024diffusionface}  DiffusionFace     & Face     & ~~6K / 120K & 11 & ~24K / 480K  & 11 & \sigyes & \signo  \\
    2024 & \cite{cheng2024diffusion}     DiFF              & Face     & 2.4K / 54K~ & 13 & 484K / 21K~  & 13 & \sigyes & \signo  \\
    2024 & \cite{zhu2024genimage}        GenImage          & Object   & ~50K / 400K &  8 & ~1.3M / 10.4M & 8 & \sigyes & \signo  \\
    2024 & \cite{park2024performance}    Park2024          & General  &  40K / 71K  & 23 & 560K / 560K  & 2  & \signo  & \signo  \\
    2024 & \cite{boychev2024imaginet}    ImagiNet          & General  &  20K / 20K  &  8 &  80K / 80K   & 8  & \sigyes & \signo  \\
    2024 & \cite{hong2024wildfake}       WildFake          & General  & 203K / 536K & 23 & 811K / 2.1M  & 23 & \signo  & \signo  \\ 
    2024 & \cite{lu2024seeing}           Fake2M            & General  & 139K / 308K & 11 & ~~1M / 2.3M  & 3  & \sigyes & \signo  \\ 
    2025 & \hspace{5mm}  \NAME~ (Ours)                     & General  &  36K / 36K  & 36 & 144K / 144K  & 36 & \sigyes & \sigyes \\
    \bottomrule
    \end{tabular}
    }
    \label{tab:sota_benchmark}
\end{table*}

Numerous methods have been proposed for the task of distinguishing AI-generated images from real ones. To understand their generalization ability, these methods are often trained on images from a single generator and then tested on images  from other generators. However, a more realistic scenario involves evaluating generalization in an online setting, where the model is trained incrementally by incorporating new generators while preserving the historical order of their release dates \cite{epstein2023online}. To address this, we present \NAME , a novel benchmark designed to evaluate and advance models capable of distinguishing AI-generated images from real ones. To ensure research reproducibility and foster innovation in the field, we will make datasets and code publicly available, and propose an evaluation protocol that facilitates the validation of new models. A key objective is to provide accessible and user-friendly tools for nonprofessionals, such as journalists, investigators, and content moderators, enabling them to easily assess the authenticity of digital information.

\NAME~focuses on high-quality, realistic images, such as those frequently shared or published on social networks, while deliberately excluding non-realistic content such as drawings, cartoons, low-resolution or noise-corrupted images. Unlike many existing benchmarks \cite{he2021forgerynet, chen2024diffusionface}, \NAME~is not limited to faces or human subjects but encompasses a broad range of visual content, thereby reflecting the diversity of real-world applications. \NAME~is not just another benchmark; it is specifically designed to address the critical limitations of existing benchmarks and datasets in the field. These limitations include arbitrary training and validation splits, which can result in biased or unreliable outcomes, unfair comparisons between methods due to inconsistent evaluation protocols, and high computational resource demand. These limitations will be discussed in detail in the following sections.

Identifying images from known generators (i.e., those used during training) is relatively straightforward. However, the real challenge lies in generalizing to images produced by unseen generators. A fundamental aspect of \NAME~is its temporal evaluation framework, where models are trained on generators from an earlier time period and validated on generators from a subsequent period. This methodology assesses the model's ability to generalize to novel generation techniques, such as the transition from GANs to diffusion models. The process is repeated by shifting the time intervals, thereby ensuring continuous adaptation to evolving models. Although this idea was introduced by Adobe researchers in \cite{epstein2023online}, there is no established benchmark for the community. Figure \ref{fig:teaser} provides a conceptual overview of the benchmark, illustrating how detection models are progressively updated and evaluated as new generators emerge. This process is explained in detail in subsequent sections.

To identify the most effective deepfake detection methods, it is essential to establish clear and fair rules and to train these methods using identical datasets and augmentation strategies. The latter can significantly affect model performance, particularly in realistic scenarios; thus, it is important that the same amount of augmentation is applied across all tested models. Furthermore, the training set is designed with scalability in mind, ensuring that a model of moderate complexity can be trained on a workstation equipped with a recent GPU in about 24 hours. This approach minimizes resource demands and enables participants with limited hardware capabilities to compete on an equal footing. In real-world applications, we expect that the best-performing models can be scaled up by retraining them on larger datasets prior to deployment.

Overall, we propose a benchmark designed to enhance the detection of AI-generated content by providing a robust, fair, and scalable framework for evaluation. By addressing the limitations of existing benchmarks, our benchmark enables researchers to develop more effective and generalizable models. Additionally, nonprofessionals will have access to simple yet effective tools for analyzing the authenticity of digital media.

\section{Related work}

In recent years, several benchmarks have been proposed for the detection of AI-generated images, accompanied by a variety of forensic detectors and datasets. In the following we will review the most relevant ones. 

\paragraph{Benchmarks}
Table ~\ref{tab:sota_benchmark} presents a comprehensive list of benchmarks for synthetic image detection. 
Some initial benchmarks only focus on GAN-based synthetic generators~\cite{he2021forgerynet} and are primarily concerned with face images, such as  
ForgeryNet~\cite{he2021forgerynet}, DiffusionFace~\cite{chen2024diffusionface}, and DIFF~\cite{cheng2024diffusion}, or on artworks~\cite{wang2023benchmarking}.
Although these analyses offer valuable insights, their applicability is currently restricted to specific image categories. 

Several generic benchmarks have been introduced with the goal of providing large-scale and diverse training datasets to enhance model generalization \cite{zhu2024genimage, hong2024wildfake, boychev2024imaginet}, or offering open-source frameworks to facilitate the integration of new models \cite{schinas2024sidbench}. Others benchmarks focus on conducting extensive analyses of publicly available datasets \cite{park2024performance}, 
including evaluations of human performance in detecting synthetic images \cite{lu2024seeing}. Additionally, several papers, while not explicitly presenting benchmarks, propose datasets that are widely adopted in the forensics community, including both GAN-based \cite{wang2020cnn} and diffusion-based images \cite{ojha2023univfd, corvi2023detection, bammey2023synthbuster, cazenavette2024fakeinversion}.

The experimental analysis conducted in the aforementioned benchmarks primarily focus on studying the generalization ability by including synthetic generators during testing that were not present during training. However, in realistic scenarios, new generators are released continuously. Therefore, it is essential to investigate generalization ability under these conditions, i.e., training on older generators and testing on newer ones based on the historical release dates of the generation models.
This type of analysis was first conducted in \cite{epstein2023online}, revealing a significant decrease in performance when generators with major architectural changes emerge. While this work demonstrated the value of such analysis, it did not establish a standardized benchmark for the research community, which is a critical gap that our work directly addresses. Furthermore, the analysis in \cite{epstein2023online} is limited to 14 generative models, while we expand this scope by incorporating 36 distinct generative models.

\paragraph{Detection Methods}
Initial approaches for distinguishing synthetic from real images primarily relied on CNN-based architectures trained on large datasets \cite{lin2024detecting}. These methods demonstrate strong performance when test conditions perfectly match the training distribution. However, they suffer from two main limitations in practical scenarios: robustness to common image impairments, such as re-compression, resizing, or cropping that frequently occur during online sharing, and generalization to unseen generative architectures \cite{Tariang2024synthetic}. To enhance robustness, a golden rule is to include carefully designed data augmentation during training. This approach not only strengthens the ability to handle image distortions but also improves generalization capabilities \cite{wang2020cnn}. Relying on pre-trained models is also very effective, especially if large pre-trained vision-language models are used. In this respect, it is possible to achieve very good performance even by relying on one single generator during training \cite{ojha2023univfd}.

Some methods propose to modify the architectures in order to better exploit low-level and/or high-level forensic traces \cite{koutlis2024rine, sarkar2024shadows} while others focus on improving the training strategy \cite{baraldi2024code,boychev2024imaginet} or by simulating the generator artifacts \cite{Rajan2024effectiveness,guillaro2024bfree}. Another path towards improving generalization is the use of few-shot or incremental learning strategies, as done in \cite{Laiti2024conditioned,li2023continual,tian2024dynamic}. These are interesting approaches, but require some images from the new models, which may not be available in the most challenging scenarios. Instead, we think that a more practical and realistic framework is to regularly re-train the detector by preserving the temporal order of the synthetic generator release date \cite{epstein2023online}.
This approach allows us to fully leverage the forensic traces of known synthetic generators which will very likely be similar to the newer generators. Indeed, it is reasonable to believe that artificial fingerprints \cite{corvi2023intriguing} from one generator likely enable classifiers to generalize across entire families of models, not just individual ones \cite{wang2020cnn}. We will experimentally show that this is the case and that major changes in the underlying generative architecture can be an issue only as soon as a completely new model is released.

\section{The benchmark}
In this section, we present the design and implementation of \NAME, a novel benchmark for evaluating synthetic media detection methods. Our framework consists of two key components: a temporally ordered dataset, organized by the chronological release dates of generative models, and a standardized evaluation and training protocol that ensures fair and reproducible comparisons. The dataset allows us to assess the generalization ability of classifiers to detect new, unseen architectures under the same training/validation split, controlled augmentation strategies, and computational constraints to simulate practical deployment conditions. 

\subsection{Dataset}
\begin{figure}
\centering
\includegraphics[width=1.0\linewidth,page=2]{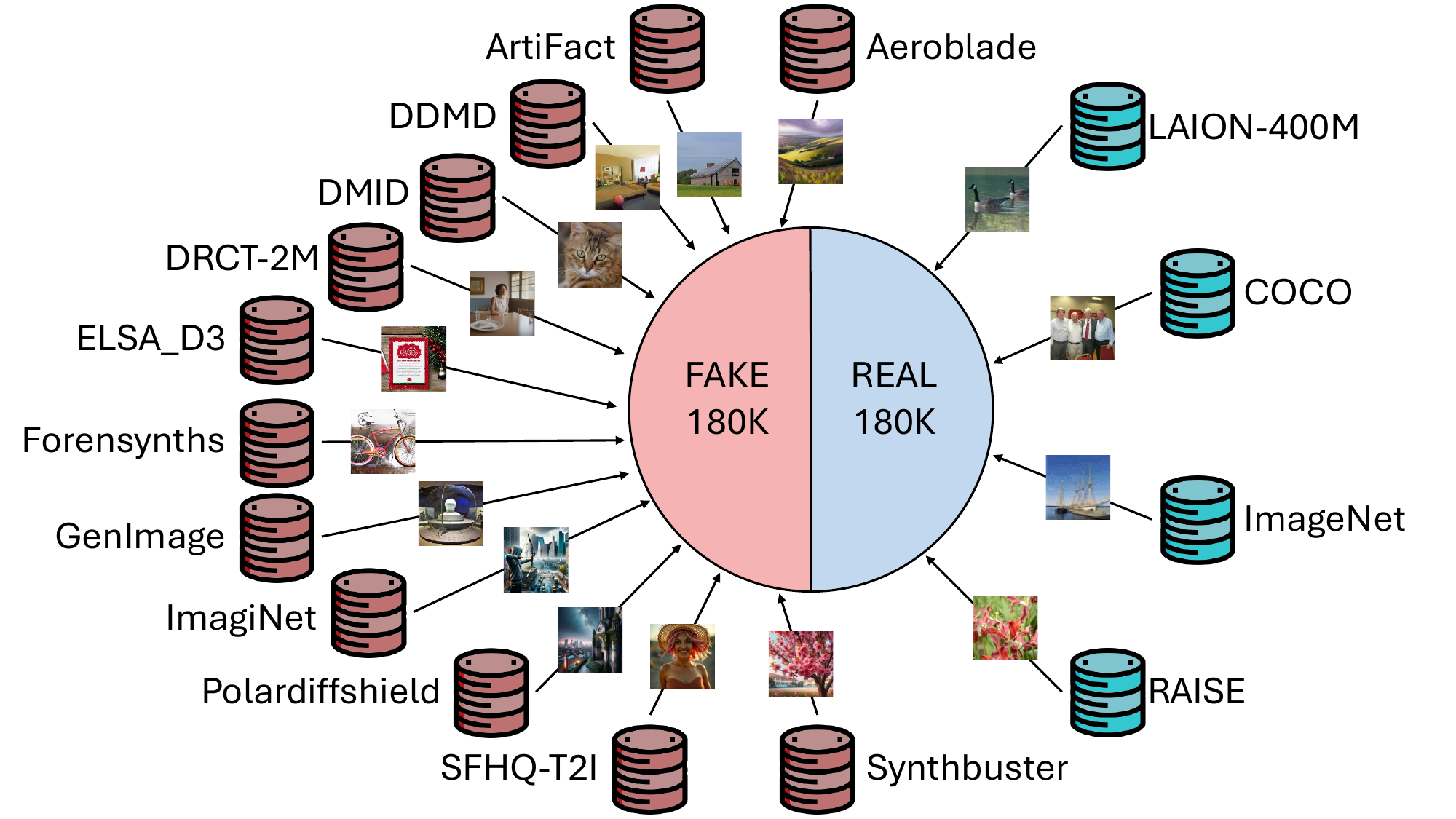}
\caption{A scheme of the content of the proposed dataset. Different data repositories of both synthetic and real images are merged to create a unified dataset containing images created using the major generative methods of the last seven years.}
\label{img:included_datasets}
\end{figure}

The dataset comprises 180K synthetic images generated by 36 different generators, each contributing 5K images. Additionally, the dataset includes 180K real images from four different sources: ImageNet (ILSVRC2012) \cite{ImageNet}, COCO2017 \cite{COCO}, LAION-400M \cite{schuhmann2021laion}, and RAISE \cite{RAISE}. To maximize diversity, synthetic images are obtained from various synthetic image repositories, listed alphabetically: Aeroblade \cite{ricker2024aeroblade}, ArtiFact \cite{rahman2023artifact}, Towards the Detection of Diffusion Model Deepfakes (DDMD) \cite{ricker2024towards}, Diffusion Model Image Detection (DMID) \cite{corvi2023detection}, DRCT-2M \cite{chen2024drct}, ELSA\_D3 \cite{baraldi2024code}, ForenSynths \cite{wang2020cnn}, SFHQ-T2I \cite{david_beniaguev_2024_SFHQ_T2I}, GenImage \cite{zhu2024genimage}, ImagiNet \cite{boychev2024imaginet}, Polardiffshield \cite{polardiffshield}, and Synthbuster \cite{bammey2023synthbuster}. A general overview of the dataset and the included repositories is presented in Figure \ref{img:included_datasets}.

We selected these data sources because they are publicly available and have been widely used in scientific research. For each generator, our dataset includes images taken from multiple sources, as these feature overlapping sets of generators. Although the original repositories have different structures and metadata, our dataset presents a unified structure (using the Arrow format). To facilitate the setup of the dataset, we provide the code used to download the files from the various sources and organize them into a common format.

The complete list of the 36 synthetic image generators $g_i, i = 0..35$ included in our dataset is reported in Table \ref{tab:benchmark_generators}. 
The dataset is divided into training and evaluation sets, following an $80\% - 20\%$ proportion, resulting in 288K images for the training set and 72K images for the evaluation set. In particular, for each generator, 4K images are used for training and 1K for evaluation. Since the dataset is balanced, real images are also split according to the same $80\% - 20\%$ proportion.

\begin{table}[t]
\groupedRowColors{4}{2}{gray!15}{white}
    \centering
    \caption{Full list of generators included in the benchmark, ordered by release date. The content of each chronological sliding window is highlighted.}
    \label{tab:benchmark_generators}
    \begin{tabular}{|c|c|l|c|}
    \hline
    \boldmath $w_j$ & \boldmath $g_i$ & \textbf{Generator} & \textbf{Release date} \\
    \hline
    & 0  & CycleGAN \cite{zhu2017unpaired} & 2017-03 \\
    & 1  & Cascaded Refinement Networks \cite{chen2017photographic} & 2017-07 \\
    & 2  & ProGAN \cite{karras2018progressive} & 2017-10 \\
    \multirow{-4}{*}{0} & 3  & StarGAN \cite{choi2018stargan} & 2017-11 \\

    & 4  & SN-PatchGAN \cite{yu2019free} & 2018-06 \\
    & 5  & BigGAN \cite{brock2018large} & 2018-09 \\
    & 6  & IMLE \cite{li2019diverse} & 2018-11 \\
    \multirow{-4}{*}{1} & 7  & StyleGAN1 \cite{karras2021style} & 2018-12 \\
    
    & 8  & GauGAN \cite{park2019semantic} & 2019-03 \\
    & 9  & StyleGAN2 \cite{karras2020analyzing} & 2019-12 \\
    & 10 & DDPM \cite{ho2020denoising} & 2020-06 \\
    \multirow{-4}{*}{2} & 11 & CIPS \cite{anokhin2021image} & 2020-11 \\
    
    & 12 & VQGAN \cite{esser2021taming} & 2020-12 \\
    & 13 & GANsformer \cite{hudson2021generative} & 2021-03 \\
    & 14 & ADM \cite{dhariwal2021diffusion} & 2021-05 \\
    \multirow{-4}{*}{3} & 15 & StyleGAN3 \cite{karras2021alias} & 2021-06 \\
     
    & 16 & LaMa \cite{suvorov2022resolution} & 2021-09 \\
    & 17 & FaceSynthetics \cite{wood2021fake} & 2021-09 \\
    & 18 & ProjectedGAN \cite{sauer2021projected} & 2021-11 \\
    \multirow{-4}{*}{4} & 19 & Palette \cite{saharia2022palette} & 2021-11 \\
    
    & 20 & VQ-Diffusion \cite{gu2022vector} & 2021-11 \\
    & 21 & Denoising Diffusion GAN \cite{xiao2022tackling} & 2021-12 \\
    & 22 & Glide \cite{nichol2022glide} & 2021-12 \\
    \multirow{-4}{*}{5} & 23 & Latent Diffusion \cite{rombach2022high} & 2021-12 \\
    
    & 24 & Midjourney \cite{midjourney} & 2022-02 \\
    & 25 & MAT \cite{li2022mat} & 2022-03 \\
    & 26 & Diffusion GAN (ProjectedGAN) \cite{wang2023diffusiongan} & 2022-06 \\
    \multirow{-4}{*}{6} & 27 & Diffusion GAN (StyleGAN2) \cite{wang2023diffusiongan} & 2022-06 \\
    
    & 28 & Stable Diffusion 1.4 \cite{rombach2022high} & 2022-08 \\
    & 29 & Stable Diffusion 1.5 \cite{rombach2022high} & 2022-10 \\
    & 30 & Stable Diffusion 2.1 \cite{rombach2022high} & 2022-12 \\
    \multirow{-4}{*}{7} & 31 & DeepFloyd IF \cite{deepfloyd} & 2023-04 \\
    
    & 32 & Stable Diffusion XL 1.0 \cite{podell2024sdxl} & 2023-07 \\
    & 33 & DALL-E 3 \cite{betker2023improving} & 2023-09 \\
    & 34 & FLUX 1 Dev \cite{flux2024} & 2024-08 \\
    \multirow{-4}{*}{8} & 35 & FLUX 1 Schnell \cite{flux2024} & 2024-08 \\
    \hline
\end{tabular}
\end{table}

\subsection{Evaluation and metrics}

To evaluate the performance of different detection methods, we assign the 36 generators to nine chronological sliding windows, denoted $w_j, j=0..8$, each containing four generators $w_j=\{g_{j \times 4+t},0\leq t\leq3\}$. The deepfake detection methods are trained progressively: at each step $k$, the model is trained on all the generators within the sliding windows $w_j, j \leq k$. This approach simulates a realistic scenario where detectors are periodically retrained to keep into account the latest technological advances. As new generators are published, the model can learn from all existing generators while future ones remain unseen.
The detector's ability to detect images from past and future generators is evaluated using various metrics. Given a detector trained at step $k$ and a set of performance indicators (e.g., accuracy, AUROC) we can evaluate its performance on different subsets of evaluation images:

\begin{itemize}
    \item \textit{Next~Period}: the evaluation set includes only the generators belonging to $w_{k+1}$. This metric is particularly important as it measures the detector's ability to generalize to unseen generators, which will become available in the near future.

    \item \textit{Past~Period}: the evaluation set includes the generators belonging to $w_j, j \le k$. This scenario evaluates how well the detector performs on the generators it has already encountered.

    \item \textit{Whole~Period}: the evaluation set includes the generators belonging to both the past and next time windows ($w_j, j \le k+1$).
\end{itemize}

For all the above metrics, a plot can be drawn (e.g. Figures \ref{img:all_next}, \ref{img:all_past} and \ref{img:all_whole}) showing the performance trend across the sliding windows. As compact ranking indicators for the leaderboard, we propose using the average Area Under Receiver Operating Characteristic (AUROC) curve and the average accuracy (across all steps) in the \textit{Next~Period} scenario. These indicators are chosen because they provide valuable insights into the detector's ability to generalize to unseen generators that are being released. 

\subsection{Training and evaluation details}
To ensure a fair comparison among methods, a standardized training and evaluation workflow has been defined. The workflow pseudocode is reported in Algorithm \ref{alg:pseudo}. In the current version of the benchmark, $T_j$ includes 4K train images from each of the 4 generators associated to $w_j$ (see Table \ref{tab:benchmark_generators}) and 4K real train images. To this purpose, the whole set of real images is initially partitioned in groups and each group is associated to a generator. Analogously, $E_j$ includes 1K evalution images from each of the generators associated to $w_j$ and 1K real evaluation images. 
To train and evaluate a new method on \NAME , the only steps that need modification are the custom training augmentation (line 11) and the method training (line 13). Hereafter, more details are provided.

\begin{algorithm}[t]
\caption{\NAME~training and evaluation workflow. In the current version, the parameters $ts$, $am$, and $n$ are set to 9, 4, and 1, respectively.} \label{alg:pseudo}
\begin{algorithmic}[1]
\State \textbf{Inputs:}
\State $T_j$ - fake and real training images of sliding win. $w_j$
\State $E_j$ - fake and real evaluation images of sliding win. $w_j$
\State $ts$ - number of training steps 
\State $am$ - augmentation multiplier for training
\State $n$ - number of epochs
\State \textbf{Body:}
\State $TS \gets \{\}$ \Comment{Training Set}
\State $ES \gets DetermAugment(E_0, 1)$ \Comment{Evaluation Set}
\For{$k = 0$ \textbf{to} $ts-1$}  \Comment{$k$-th training step}
    \State $TA \gets CustomAugment(T_k, am)$ \Comment{Augm. Train}
    \State $TS \gets TS \cup TA$ \Comment{Current Training Set}
    \State $m_k \gets$ \text{Model trained} \text{for} $n$ \text{epochs on} $TS$
    \State \text{Evaluate model} $m_k$ \text{on} $ES$  \Comment{\textit{Past~Period}}
    \If{$k<ts-1$}
        \State $EN \gets DetermAugment(E_{k+1}, 1)$ \Comment{Next Eval.}
        \State \text{Evaluate model} $m_k$ \text{on} $EN$  \Comment{\textit{Next~Period}}
        \State $ES \gets ES \cup EN$ \Comment{Whole Eval. Set}
        \State \text{Evaluate model} $m_k$ \text{on} $ES$  \Comment{\textit{Whole~Period}}
    \EndIf
\EndFor
\end{algorithmic}
\end{algorithm}

\subsubsection{Augmentation}
the training augmentation pipeline (function $CustomAugment$) can be customized by the method itself, as this is often a key distinguishing feature. However, the number of (diverse) augmented images presented to the model is fixed and determined by the multiplier $am$. In this version of the benchmark, we use $am = 4$, meaning that each training image is replaced with four augmented versions of itself. This rule is enforced to ensure a fair evaluation with respect to the computational resources: by limiting the amount of augmentation allowed during training, the focus is placed on the effectiveness of the detection method itself. Without this limitation, methods relying on more computational resources could achieve better results by leveraging more extensive augmentations. 
In contrast, the evaluation augmentation (function $DetermAugment$) is deterministic and cannot be customized (here the multiplier is 1).
The goal of the augmentation pipeline is to produce images that closely resemble those typically published on media platforms or shared via instant messaging applications.
The pipeline includes a combination of compression, blurring, noise addition, variable resizing, and cropping steps.
These transformations disrupt most of the noise patterns found in generated images while still producing good quality images that could reasonably be accepted and re-shared by human users.

\subsubsection{Additional data and pretrained models}
training on additional data or using models pre-trained on the same problem is not allowed. However, general-purpose open-weight foundation models can be used as backbones.

\subsubsection{Supporting codebase, leaderboard and reproducibility}
a codebase will be released to facilitate the download and setup of the dataset, as well as the execution of the training and evaluation workflow described in Algorithm \ref{alg:pseudo}. A plug-in architecture has been designed to simplify the addition of new methods through the customization of a few functions. The codebase also includes baseline methods for comparison and a default training augmentation pipeline.
A leaderboard will be maintained for each version of the benchmark. Methods added to the leaderboard must include an accompanying white paper and scripts, ensuring full reproducibility of the results.

\begin{figure}[t]
\centering
\includegraphics[width=1.0\linewidth,page=1]{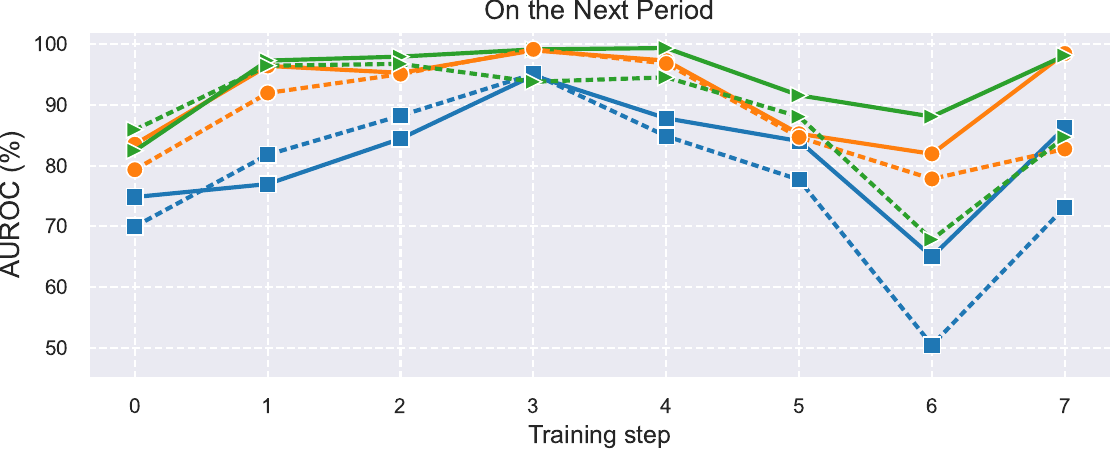}
\includegraphics[width=1.0\linewidth,page=2]{figures/results_next.pdf}
\includegraphics[width=1.0\linewidth,page=3]{figures/results_next.pdf}
\caption{AUROC and Accuracy for different models and training strategies on generators in the \textit{Next~Period} on resized images.}
\label{img:all_next}
\end{figure}

\begin{figure}[t]
\centering
\includegraphics[width=1.0\linewidth,page=1]{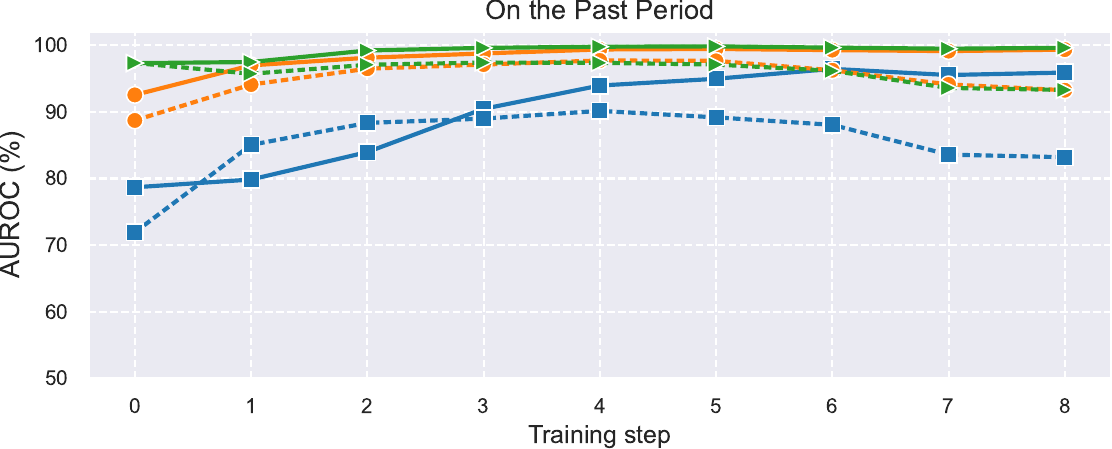}
\includegraphics[width=1.0\linewidth,page=2]{figures/results_past.pdf}
\includegraphics[width=1.0\linewidth,page=3]{figures/results_past.pdf}
\caption{AUROC and Accuracy for different models and training strategies on generators in the \textit{Past~Period} on resized images.}
\label{img:all_past}
\end{figure}

\begin{table}[t]
    \setlength{\tabcolsep}{4.5pt}
    \centering
    \caption{Comparison of the average AUROC and Accuracy for Resize and Multicrop Experiments on the Next~Period.}
    \label{tab:auroc_accuracy}
    \begin{tabular}{lc|cc|cc}
        \toprule
        \multirow{2}{*}{\textbf{Model}} & \multirow{2}{*}{\textbf{Mode}} & \multicolumn{2}{c|}{\textbf{AUROC}} & \multicolumn{2}{c}{\textbf{Accuracy}} \\
        \cmidrule(lr){3-4} \cmidrule(lr){5-6}
        &  & \textbf{Resize} & \textbf{Multicrop} & \textbf{Resize} & \textbf{Multicrop} \\
        \midrule
        ResNet-50 CLIP        & Tune  & 81.77  & 76.51  & 73.42  & 67.22  \\
        ResNet-50 CLIP        & Probe & 77.66  & 69.08  & 69.85  & 63.33  \\
        ViT-L/14 CLIP        & Tune  & 92.04  & 87.66  & \textbf{85.28}  & 76.83  \\
        ViT-L/14 CLIP        & Probe & 88.47  & 68.74  & 76.25  & 63.02  \\
        ViT-L/14 DINOv2      & Tune  & \textbf{94.24}  & \textbf{93.44}  & 84.09  & \textbf{78.64}  \\
        ViT-L/14 DINOv2      & Probe & 88.47  & 88.74  & 79.34  & 78.09  \\
        \bottomrule
    \end{tabular}
\end{table}

\section{Performance of baseline methods}

We trained and evaluated a set of baseline methods using the workflow described in Algorithm \ref{alg:pseudo}. These methods are based on well-known pre-trained vision models: i) \textit{ResNet-50 CLIP} by OpenAI \cite{radford2021learning},  ii) \textit{ViT-L/14 CLIP} from LAION models\footnote{\href{https://huggingface.co/laion/CLIP-ViT-L-14-CommonPool.XL-s13B-b90K}{laion/CLIP-ViT-L-14-CommonPool.XL-s13B-b90K}}, and iii) \textit{ViT-L/14 DINOv2} \cite{oquab2024dinov}. The input size for all the models is 224x224. For each model, we explored two distinct training strategies: \textit{fine-tuning} and \textit{linear probing}. The fine-tuning strategy (hereafter referred to as \textit{tune}) involves training the entire model, while the linear probing strategy (hereafter referred to as \textit{probe}) involves training only the final classification layer built on top of a frozen backbone. In both cases, at each training step, the models have been trained for a single epoch ($n=1$).

Additionally, we investigated two approaches for converting images of variable sizes into the fixed-size input images required by the used models: i) \textit{resize}, where images are resized to a fixed size during both training and evaluation; and ii) \textit{multi-cropping}, where a single random crop is used during training, while five crops (the center and the four corners) are employed during evaluation, with their predictions aggregated via averaging. The resize strategy better preserves the semantic content of the images, at the cost of altered resolution and aspect ratio, which may attenuate or remove artifacts introduced by generative methods. In contrast, the multi-cropping strategy analyzes the images at their original resolution, at the cost of using only partial content information.  This results in a total of 12 experiments: 3 models $\times$ 2 training modes $\times$ 2 image adaptation mechanisms.
At the start of the training process (line 13 of Algorithm \ref{alg:pseudo}), for each step $k>0$, we reload the weights from the model trained on the previous step $m_{k-1}$, rather than reloading weights from the original pre-training. This approach simulates an incremental improvement of the detection model over time.

\begin{figure}[t]
\centering
\includegraphics[width=1.0\linewidth,page=1]{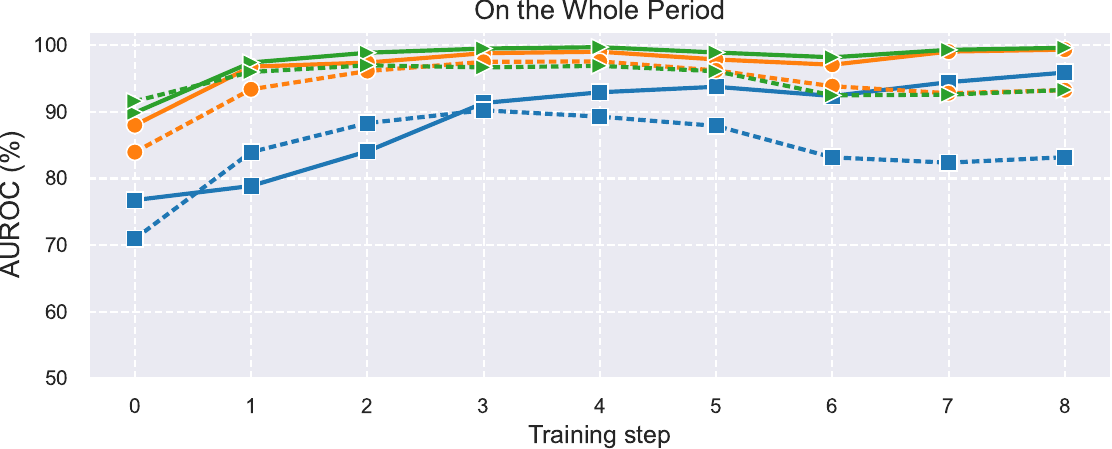}
\includegraphics[width=1.0\linewidth,page=2]{figures/results_whole.pdf}
\includegraphics[width=1.0\linewidth,page=3]{figures/results_whole.pdf}
\caption{AUROC and Accuracy for different models and training strategies on generators in the \textit{Whole~Period} on resized images.}
\label{img:all_whole}
\end{figure}

Figures \ref{img:all_next}, \ref{img:all_past}, and \ref{img:all_whole} show the AUROC and Accuracy trends when using image resizing for \textit{Next~Period}, \textit{Past~Period} and \textit{Whole~Period} scenarios, respectively.
It is evident that \textit{Next~Period} scenario is more challenging since detectors are evaluated only on unseen generators and technological changes can determine significant accuracy drops. On the contrary, in the \textit{Whole~Period} scenario, the increasing weight of known generation techniques leads to a more stable improvement.

To be more specific, in Figure \ref{img:all_next}, two noticeable performance drops are observed. A significant drop occurs at step 6 when detectors trained on $w_i, i=0 \ldots 6$ are assessed on $w_7$. Referring to Table \ref{tab:benchmark_generators}, this drop coincides with the model's first evaluation on the \textit{Stable Diffusion (1.4-2.1 $+$ DF-IF)} generators, which is a significant milestone in the evolution of image generation techniques. A minor drop is observed at step 4 (training on $w_i, i=0 \ldots 4$ and evaluation on $w_5$), corresponding to the introduction of \textit{VQ-Diffusion}, \textit{Denoising Diffusion GAN}, \textit{Glide}, and \textit{Latent Diffusion}, which are precursors to the next generation of robust diffusion techniques.

\begin{figure}[t]
\centering
\includegraphics[width=1.0\linewidth,page=1]{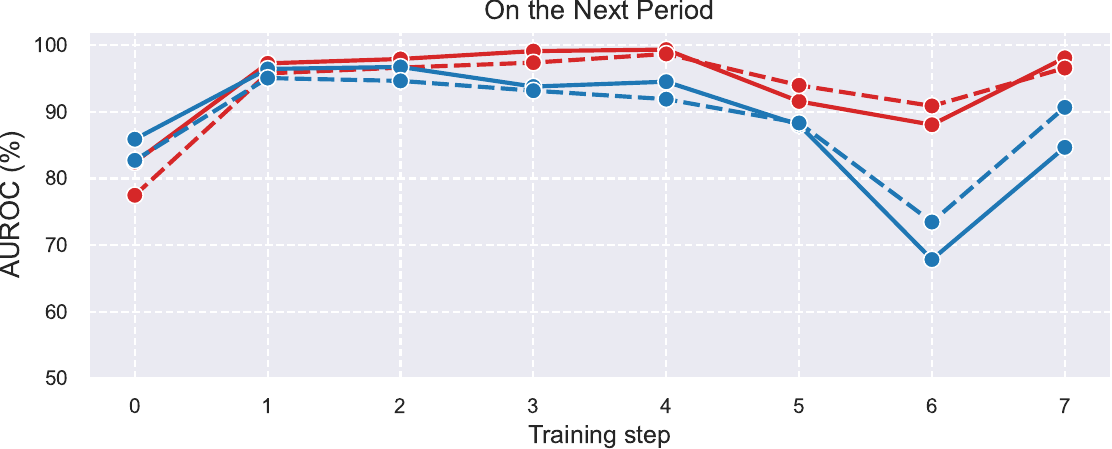}
\includegraphics[width=1.0\linewidth,page=2]{figures/results_d_next.pdf}
\includegraphics[width=1.0\linewidth,page=3]{figures/results_d_next.pdf}
\caption{AUROC and Accuracy for the \textit{ViT-L/14 DINOv2} variants on generators in the \textit{Next~Period}.}
\label{img:dino_next}
\end{figure}

If the focus is moved to the comparison of the different baseline methods, besides Figure \ref{img:all_next} and \ref{img:all_whole} we can consider Table \ref {tab:auroc_accuracy} reporting the average AUROC and Accuracy. We observe that: i) larger models outperform smaller ones; ii) fine-tuning the entire backbone consistently leads to better generalization compared to freezing it during training; iii) resize is often preferable with respect to multi-crop.
Specific details for \textit{ViT-L/14 DINOv2} variants are provided in Figure \ref{img:dino_next}.

Table \ref{tab:training_time} reports the training times of the baseline methods on a workstation with a single GPU, showing that the entire training process (following the Algorithm \ref{alg:pseudo} workflow) of the most resource-demanding model can be completed in about 1 day. For smaller models, such as the ResNet-50, data loading and augmentation (executed on the CPU) create a performance bottleneck, resulting in equal times for tune and probe.

\begin{table}[ht]
    \setlength{\tabcolsep}{4.5pt}
    \centering
    \caption{Training times of baselines according to Algorithm \ref{alg:pseudo}. The reference system is equipped with an \textit{Intel i9-10900X} CPU and a \textit{NVIDIA GeForce RTX 3080 Ti} GPU.}
    \label{tab:training_time}
    \begin{tabular}{lc|c}
        \toprule
        \textbf{Model} & \textbf{Mode} & \textbf{Training Time} \\
        \midrule
        ResNet-50 CLIP       & Tune  & 4.87h \\
        ResNet-50 CLIP       & Probe & 4.87h \\
        ViT-L/14 CLIP        & Tune  & 19.87h \\
        ViT-L/14 CLIP        & Probe & 6.35h \\
        ViT-L/14 DINOv2      & Tune  & 26h  \\
        ViT-L/14 DINOv2      & Probe & 7.5h  \\
        \bottomrule
    \end{tabular}
\end{table}

\section{Conclusions}
\NAME~is introduced as a novel, ongoing benchmark for detecting AI-generated images in real-world scenarios. This benchmark will provide datasets, tools, and leaderboards to researchers and practitioners, facilitating the training, evaluation, and fair comparison of their detection methods. As we write this paper, new generation techniques are rapidly gaining popularity, and we plan to incorporate these techniques in future versions of the benchmark, including additional temporal sliding windows to keep the evaluation up-to-date with the latest technological advancements. In the future, we aim to consider local alteration and generation techniques, such as image inpainting, and extend the evaluation framework to include videos.
The baseline methods presented in this paper are based on simplified design choices, which we intend to reconsider and refine. This reassessment will help determine if these design choices can benefit existing detectors, thereby distilling guidelines that may support the development of novel approaches.

\ifCLASSOPTIONpeerreview
\else
\section*{Acknowledgment}

We acknowledge, for the first author, the support of the European funds from the Emilia-Romagna Region under the Fse+ 2021-2027 programme.
In addition, this work has received funding from the European Union under the Horizon Europe vera.ai project, Grant Agreement number 101070093, and was partially supported by SERICS (PE00000014) under the MUR National Recovery and Resilience Plan, funded by the European Union - NextGenerationEU. 
\fi

\printbibliography

\end{document}